\documentclass{article}

\usepackage{arxiv}

\usepackage[utf8]{inputenc} 
\usepackage[T1]{fontenc}    
\usepackage{hyperref}       
\usepackage{url}            
\usepackage{booktabs}       
\usepackage{amsfonts}       
\usepackage{nicefrac}       
\usepackage{microtype}      
\usepackage{lipsum}
\usepackage{multirow} 
\usepackage{float}
\usepackage{amsmath}
\usepackage{graphicx}
\graphicspath{ {./images/} }

\title{Leveraging Scene Geometry and Depth Information for Robust Image Deraining}

\author{
 Ningning Xu and Jidong J. Yang$^{}$*\\
  Smart Mobility and Infrastructure Lab\\
  College of Engineering\\
  University of Georgia\\
  Athens, GA 30602 \\
  \texttt{(Ningning.Xu@uga.edu; Jidong.Yang@uga.edu) } \\
}

\begin{document}
\maketitle
\begin{abstract}
Image deraining holds great potential for enhancing the vision of autonomous vehicles in rainy conditions, contributing to safer driving. Previous works have primarily focused on employing a single network architecture to generate derained images. However, they often fail to fully exploit the rich prior knowledge embedded in the scenes. Particularly, most methods overlook the depth information that can provide valuable context about scene geometry and guide more robust deraining. In this work, we introduce a novel learning framework that integrates multiple networks: an AutoEncoder for deraining, an auxiliary network to incorporate depth information, and two supervision networks to enforce feature consistency between rainy and clear scenes. This multi-network design enables our model to effectively capture the underlying scene structure, producing clearer and more accurately derained images, leading to improved object detection for autonomous vehicles. Extensive experiments on three widely-used datasets demonstrated the effectiveness of our proposed method.
\end{abstract}


\section{Introduction}
Image deraining is a critical preprocessing step in computer vision applications due to its significant impact on visual clarity and accuracy. Rain on images can obscure the visibility of objects, leading to substantial degradation in image quality. This can adversely affect the performance of object detection\cite{zhao2019object}, recognition\cite{logothetis1996visual}, and tracking algorithms\cite{kang2011automatic}, which are essential in various domains such as surveillance and navigation. In autonomous driving\cite{yurtsever2020survey}, clear vision is paramount for safety and robust decision-making; rain-induced artifacts can compromise the accuracy of perception systems, potentially leading to hazardous situations. Therefore, effective image deraining techniques are vital to enhance the reliability and functionality of vision-based systems.

In general, a rainy image can be represented as a superimposition of two layers: a clean image layer and a rain layer.  The rain layer encompasses various artifacts such as rain streaks, raindrops, and fog, which make rain removal a challenging task. These rain-induced artifacts obscure objects and scenes, not only blurring visual data but also partially concealing critical features necessary for accurate image interpretation. Moreover, spatial factors further complicate the process. For objects are closer to the camera, rain is the primary occluding element, making its removal relatively easier.  In contrast, distant objects are more challenging to recover due to additional occlusions from fog and other atmospheric conditions. This intricacy underscores the challenges of deraining in computer vision.  Garg and Nayar\cite{garg2007vision} illustrated this phenomenon by demonstrating how the intensity of rain effects transitions into fog as the distance to the scene increases. Recent deep learning approaches for single-image rain removal have predominantly concentrated on the removal of rain streaks, often neglecting the broader physical characteristics of rain itself. The existing training datasets for rain removal typically include images featuring artificial rain streaks, raindrops, or a combination of both, with some datasets even containing indoor scenes. This limitation significantly hampers the effectiveness of these methods when applied to real-world outdoor scenarios, where the intricate interplay between rain patterns and environmental factors differs substantially from the synthetic conditions represented in these datasets.

In this study, we proposes a novel method for automatic removal of rain streaks, raindrops, and fog in real-world photographs, with an emphasis on achieving real-time performance. The primary objective is to improve image quality for environmental monitoring and vision-based autonomous driving, thereby enhancing the accuracy and reliability of these applications under challenging weather conditions. To achieve this, we introduce an autoencoder model equipped with a consistent feature extraction module that processes both rainy and clear images while incorporating depth information. This approach allows the model to capture the underlying shared features between rain and clear images, thereby preserving the essential scene information obscured by rain and fog. Furthermore, the integration of depth information  enables the extraction of global image features, ensuring the retention of key structural details across entire images.

In summary, this work has the following contributions:
\begin{itemize}
    \item Firstly, we constructed a Derain AutoEncoder (DerainAE) model to effectively handle various rain-related artifacts and atmospheric disturbances.
    \item Secondly, we designed a consistent feature extraction module with a supervision network during training to effectively capture shared features between rain and clear images. 
    \item  Thirdly, we developed a depth network (DepthNet) to extract depth information, which aids in capturing global structure of scenes. By leveraging these shared and global features, our deraining model is capable of generating more accurate and visually coherent results.
    \item Lastly, we conducted extensive experiments to evaluate our approach on various outdoor datasets. The results showed that our method effectively removes rain artifacts while preserving critical image details. The efficacy of our model was further validated through its performance on an object detection task.
\end{itemize}

\section{Related Work}

\subsection{Image Deraining}
Image deraining methods can be broadly categorized into model-based methods\cite{chen2013generalized,gu2017joint,luo2015removing} and deep learning methods\cite{zhang2022beyond,ren2019progressive,chen2024bidirectional,hou2024robustness}. Model-based methods often approach deraining as a filtering problem, using various filters to restore a rain-free image. While this can effectively remove rain effects, it also tends to eliminate important structures within the image. Many model-based approaches develop various image priors based on the statistical properties of rain and clear images. These methods include image decomposition\cite{kang2011automatic}, low-rank representation\cite{chen2013generalized,zhang2017convolutional}, discriminative sparse coding\cite{zhu2017joint}, and Gaussian mixture models\cite{li2016rain}. Although these techniques have achieved improved results, they still struggle to effectively handle complex and varying rainy conditions.

In contrast, deep learning-based methods have significantly advanced image deraining by learning data-driven representations of rain and clear images. These approaches utilize powerful architectures and novel mechanisms to achieve superior performance. Early works such as \cite{fu2017clearing} demonstrated substantial improvements in rain removal across  benchmark datasets using  convolutional neural networks (CNNs).  Generative adversarial networks (GANs) \cite{goodfellow2020generative} have also been employed to restore perceptually superior rain-free images, as demonstrated by \cite{yadav2021deraingan}.  The introduction of transformers, such as \cite{zamir2022restormer}, enabled effective modeling of non-local dependencies, further enhancing image reconstruction quality.  Inspired by the success of recent diffusion models\cite{rombach2022high} in generating high-quality images, diffusion-based approaches\cite{liu2024residual} have shown great potential in improving image deraining performance across complex scenarios. Recent advancements include the integration of additional data modalities and novel priors into the learning process. For instance, Hu et al. \cite{hu2019depth} introduced depth information via an attention mechanism, achieving promising results on synthetic rainy datasets. Zhang et al. \cite{zhang2022beyond} exploited both stereo images and semantic information for improved image deraining performance. Guo et al.  \cite{guo2022exploring} proposed the use of Fourier priors to improve model generalization in rain removal tasks.

In summary, model-based methods have historically provided a solid foundation for image deraining, emphasizing handcrafted priors and optimization frameworks. However, their limitations in handling complex rainy conditions and preserving image details have led to a growing focus on deep learning approaches. Deep learning methods, driven by CNNs, GANs, transformers, and diffusion models, continue to achieve state-of-the-art results by leveraging large datasets, powerful architectures, and innovative priors. With the rapid evolution of data-driven techniques, deep learning is poised to dominate future advancements in image deraining, offering scalable solutions for complex and diverse real-world scenarios.

\section{Methods}
In this section, we introduce our multi-network approach for effective image deraining. The core of this framework is the Deraining AutoEncoder (DerainAE), which serves as the primary network for the deraining task. To enhance its performance, we introduce a supplementary Depth Network (DepthNet) that integrates depth information to assist in rain removal. Additionally, we utilize pretrained networks to provide supervisory signals, ensuring multiscale feature consistencies between clear and rainy images. The detailed architecture and loss functions are discussed subsequently.

\subsection{Network architecture}
\begin{figure}[ht]
\begin{center}
\includegraphics[width=1.0\linewidth]{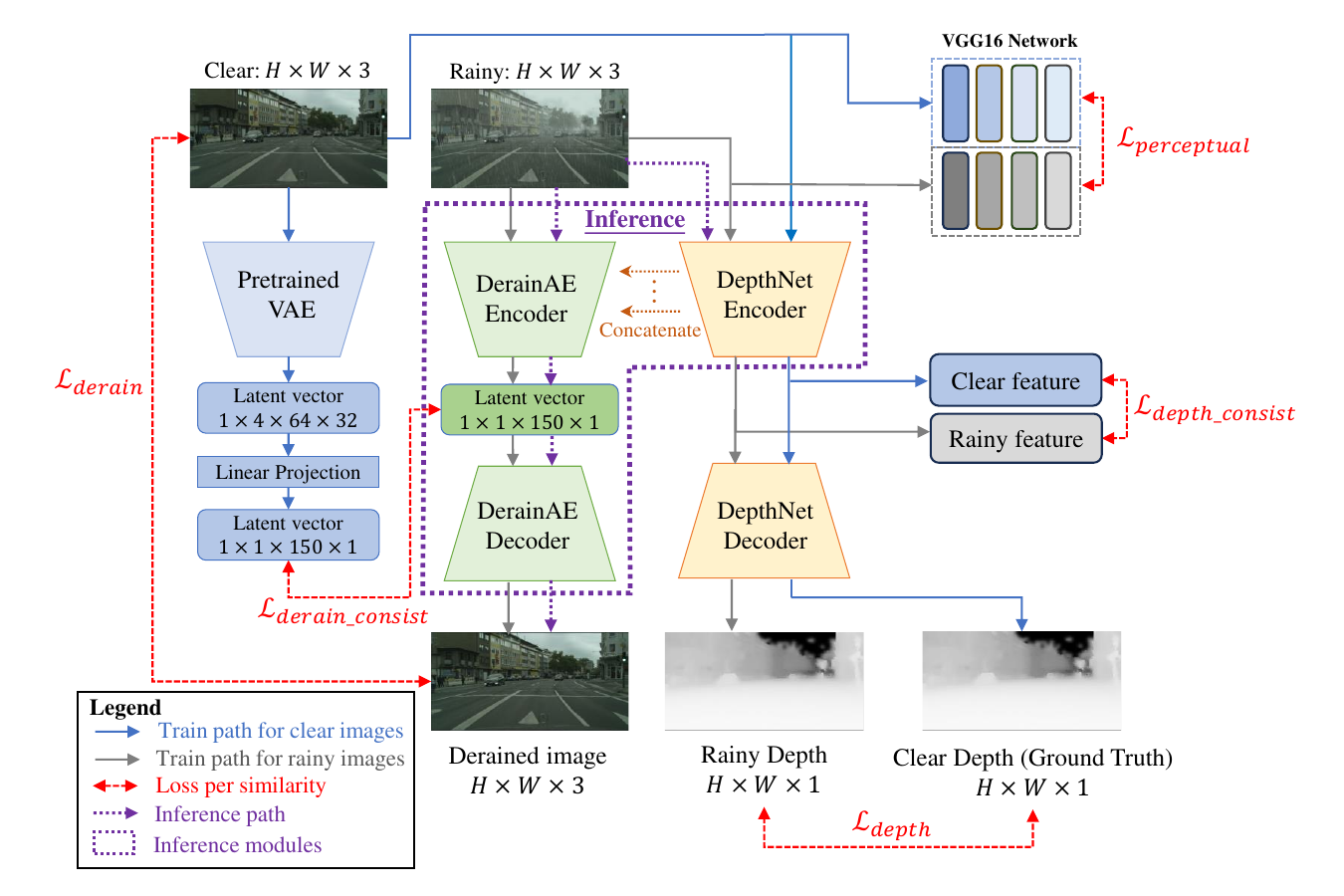}
\caption{The overall architecture of our model. A pretrained VAE extracts clear features, while the DerainAE and DepthNet modules handle rainy images. Latent space comparison between clear and rainy features improves depth estimation and deraining images prediction.}
\label{Network}
\end{center}
\end{figure}
For image deraining, the commonality between clear and rainy images lies in their depiction of the same scene, meaning the depth map should remain consistent between them. Apart from the rain artifacts, the feature map of the clear and rainy images should also be identical. Therefore, our approach employs two forms of supervision: one from the depth map and one from the feature map. This dual supervision ensures that the model not only learns to remove the rain but also retains the intrinsic features of the scene, leveraging the consistency between the depth and feature information to enhance the deraining process.

Our DerainAE model adopts an autoencoder architecture to tackle the image deraining task by learning both the latent representation and the restored derained image. The autoencoder is designed to effectively capture the underlying structure and intrinsic features of rain-affected images through an encoding-decoding process. The encoder compresses the input image into a lower-dimensional latent space, extracting critical high-level information necessary for rain removal while filtering out irrelevant noise. The decoder then reconstructs the derained image from this latent representation, ensuring the preservation of essential details and textures. This dual functionality enables the model to efficiently map rain-degraded images to their rain-free counterparts.

To enhance the learning capacity of the DerainAE model and enable it to capture more comprehensive scene information, we integrate a DepthNet that also adopts an encoder-decoder architecture.  Features from the DepthNet encoder are concatenated with the corresponding feature levels of the DerainAE encoder, establishing a shared learning mechanism that effectively leverages depth information for improved deraining performance. In our implementation, the DepthNet encoder employs the VGG16 architecture,  allowing the model to leverage depth information to better understand the spatial structure and geometry of the scene, which is crucial for accurate rain removal. The decoder employs transposed convolutions to progressively upsample the feature maps, restoring them to the original input resolution. To preserve high-resolution details, skip connections are implemented between the encoder’s convolutional blocks and their corresponding layers in the decoder, following the design principles of the U-Net architecture. Additionally, the decoder incorporates multiple convolutional layers to effectively fuse information across different spatial resolutions. The network predicts disparity maps at multiple scales and resolutions using convolutional layers with sigmoid activation functions.

During training, we use the DerainAE for image deraining while simultaneously leveraging the DepthNet to predict the depth maps of both clear and rainy images. The feature maps from DepthNet encoder are concatenated with the corresponding feature maps in the DerainAE encoder, enabling depth-aware deraining. Additionally, a pretrained Variational Autoencoder (VAE) is used to obtain a latent vector of the clear image, which serves as a supervisory signal during during training to ensure high-level feature consistency.  Feature consistency is further  enforced at multiple levels via a pretrained VGG16.  Depth consistency is also maintained in the latent space of the DepthNet.  During inference, our method requires only the rainy image as input, which is processed by DerainAE and the DepthNet encoder to produce the derained output, where the DepthNet encoder extracts depth information, which is then passed to the DrainAE encoder to aid in the deraining process.

\begin{figure}[ht]
\begin{center}
\includegraphics[width=1.0\linewidth]{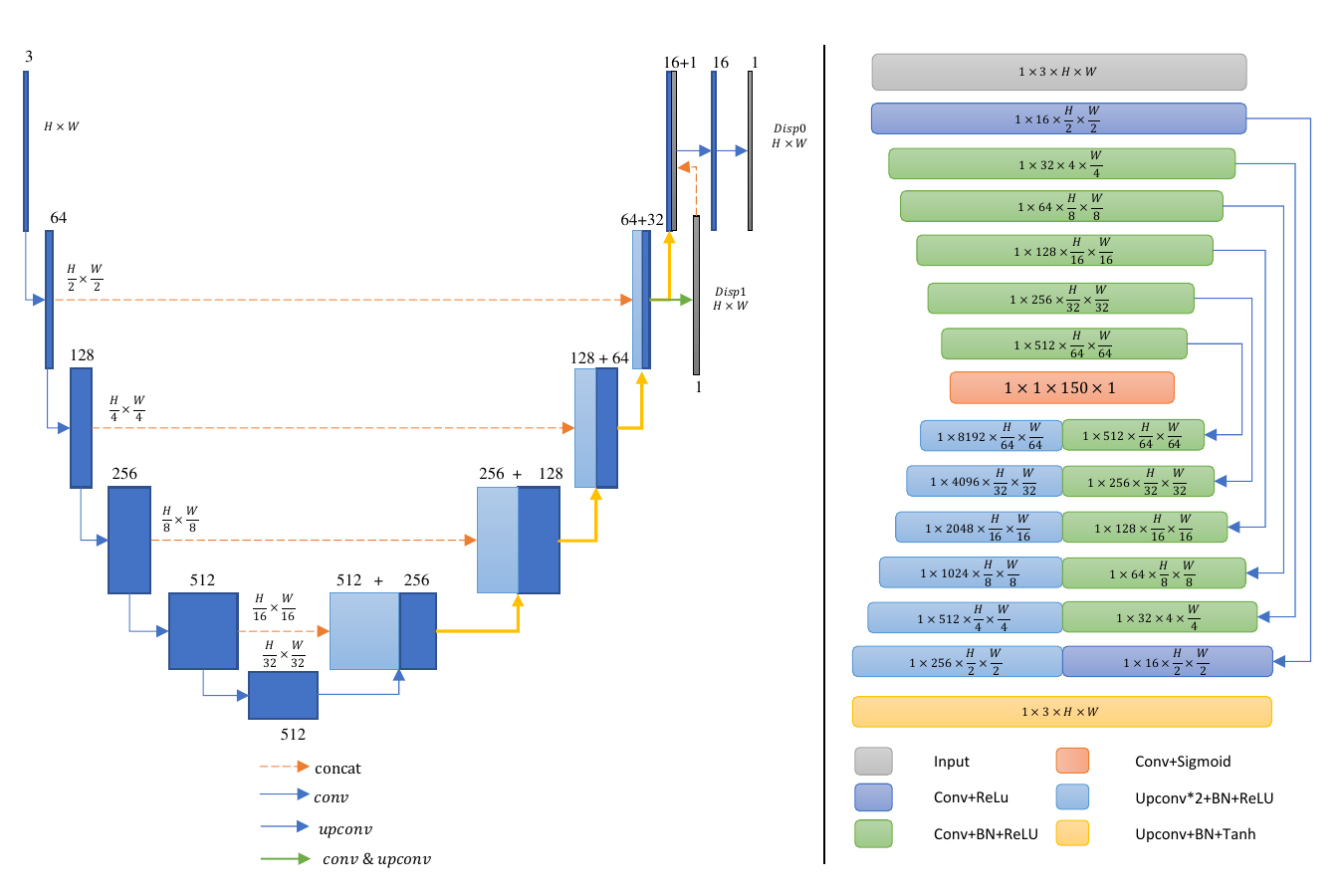}
\caption{An overview of our DepthNet and DerainAE architecture. Left: DepthNet, this model employs a U-Net structure, with skip connections from each encoder layer to the corresponding decoder layers. The network outputs two disparity maps, with Disp0 used as the final predicted depth map. Right: DerainAE, this model is a simple convolutional network with skip connections at corresponding feature levels between encoder and decoder.}
\label{DepthNet}
\end{center}
\end{figure}

\subsection{Loss function}
To jointly train DerainAE and DepthNet, we introduce a composite loss function that considers multiple complementary loss components. Building on the perceptual loss $\mathcal{L}_{\text{perceptual}}$ proposed by Johnson et al. \cite{johnson2016perceptual}, we measure the discrepancy between clear images and corresponding rain images in a manner more consistent with human visual perception. Specifically, we utilize a pretrained VGG16 network to capture discrepancies at various feature levels, computed by Equation 1.

\begin{align*} 
\mathcal{L}_{perceptual} = \sum_l \lambda_l \cdot \left| \phi_l(y) - \phi_l(\hat{y}) \right|_2^2 \tag{1}
\end{align*}

where $\phi_l$ denotes the activation map of the $l$-th layer in VGG16.

We employ cosine similarity losses (Equations 2 and 3) to measure the consistency of latent representations of clear images and corresponding rain images for both  DerainAE and DepthNet.

\begin{align*} 
\mathcal{L}_{depth\_consist} = \cos\left(D_R, D_C\right)\tag{2}\\
\mathcal{L}_{derain\_consist} = \cos\left(R_L, C_L\right)\tag{3} 
\end{align*}

where $\cos(\cdot, \cdot)$ denotes the cosine similarity function.

Additionally, we use mean squared error (MSE) losses for reconstruction of the depth map $\hat{D}$ and the derained image $\hat{C}$:

\begin{align*} 
\mathcal{L}_{derain} = MSE(\hat{C}, C)\tag{4}\\ 
\mathcal{L}_{depth} = MSE(\hat{D}, D)\tag{5}
\end{align*}

The loss function used to train our model is a weighted sum of these individual loss terms by Equation (6):

\begin{align*} 
\mathcal{L} = \lambda_1 \mathcal{L}_{perceptual} + \lambda_2 \mathcal{L}_{depth\_consist} + \lambda_3 \mathcal{L}_{derain\_consist} + \lambda_4 \mathcal{L}_{derain} + \lambda_5 \mathcal{L}_{depth}\tag{6}
\end{align*}

where $\lambda_1, \ldots, \lambda_5$ are hyperparameters that control the relative importance of each loss component during training. This hybrid loss function enables the joint optimization of DerainAE and DepthNet, ensuring robust performance across both deraining and depth estimation tasks.

\section{Experimental results}
In this section, we begin by introducing the datasets and evaluation metrics, followed by a discussion of the implementation details and results. Ablation studies are conducted to evaluate the contributions of key components. Additionally, the effectiveness of our model is validated through an object detection task, highlighting the benefits of deraining for enhanced vision.

\subsection{Datasets and evaluation metrics}
Due to the challenge of obtaining paired rain and clear images, various rain models have been developed to synthetically generate rain streaks from clear images. In the linear model proposed by \cite{li2016rain}, the observed rain image $\mathbf{O} \in \mathbb{R}^{M \times N}$ is represented as a combination of a desired background layer $\mathbf{B} \in \mathbb{R}^{M \times N}$ and a rain streak layer $\mathbf{R} \in \mathbb{R}^{M \times N}$, such that $\mathbf{O} = \mathbf{B} + \mathbf{R}$. Building upon this model, \cite{yang2017deep} proposed a more generalized formulation: $\mathbf{O} = \mathbf{B} + \mathbf{R}\mathbf{\tilde{R}}$, where $\mathbf{\tilde{R}}$ is a new region-dependent variable that indicates the locations of individually visible rain streaks. The elements of $\mathbf{\tilde{R}}$ take binary values, with 1 indicating rain regions and 0 indicating non-rain regions. Further, \cite{hu2021single, hu2019depth} modeled a rain image as a composite of a rain-free image, a rain layer, and a fog layer, formulating the observed rain image as below,
\begin{align*}
    \mathbf{O}=\mathbf{B}(1-\mathbf{R}-\mathbf{F})+\mathbf{R}+f_0\mathbf{F}
\end{align*}
where $\mathbf{F}$ is a fog layer, $f_0$ is the atmospheric light, which is assumed to be a global constant following\cite{sakaridis2018semantic}.

\textbf{RainKITTI2012 and RainKITTI2015 Datasets}: These two synthetic datasets were created by Zhang et al. \cite{zhang2022beyond} using Photoshop to introduce synthetic rain effects on the publicly available KITTI stereo datasets 2012 and 2015 \cite{Geiger2012CVPR}. The RainKITTI2012 dataset consists of a training set with 4,062 image pairs and a testing set with 4,085 image pairs, each having a resolution of $1242 \times 375$ pixels. Similarly, the RainKITTI2015 dataset contains 4,200 pairs of training images and 4,189 pairs of testing images, all maintaining the same resolution.

\textbf{RainCityScapes Dataset}: This synthetic dataset, developed by Hu et al. \cite{hu2019depth}, is based on the Cityscapes dataset \cite{cordts2016cityscapes}. It is generated by the aforementioned rain models and consists of a rain layer, a fog layer, and a rain-free image. It includes a training set of 9,432 paired rainy and clear images, accompanied by ground truth depth information. The testing set consists of 1,188 images.

\textbf{Evaluation metrics} We use PSNR\cite{huynh2008scope} and SSIM\cite{wang2004image} as the evaluation metrics.

\subsection{Implementation details}
In model training, we set the hyperparameters $\lambda_1, \lambda_2, \lambda_3, \lambda_4, \lambda_5$ to [1, 0.5, 0.5, 10, 2], respectively. For the pretrained Variational Autoencoder (VAE) model, we adopt the VAE component from the Stable Diffusion framework \cite{rombach2022high}, employing Mean Squared Error (MSE) as the loss function. The latent space of the VAE is sampled to produce a latent vector of the same size (length 150) as that used in our DerainAE model. During training, we keep the VAE model weights frozen and only fine-tune the final output layer. For depth reconstruction, we use the pretrained VGG16 model as the encoder, which is frozen during training,  and train the decoder from scratch. Our entire model is implemented in PyTorch \cite{paszke2017automatic} and is trained on a workstation with a NVIDIA RTX A6000 GPU. All datasets in our experiments share the same training configuration: a batch size of 4, and the ADAM optimizer \cite{kingma2014adam} with an initial learning rate of $5 \times 10^{-3}$ and a weight decay of 0.9.

\subsection{Evaluation on Different Datasets}
\begin{figure}[ht]
\begin{center}
\includegraphics[width=1.0\linewidth]{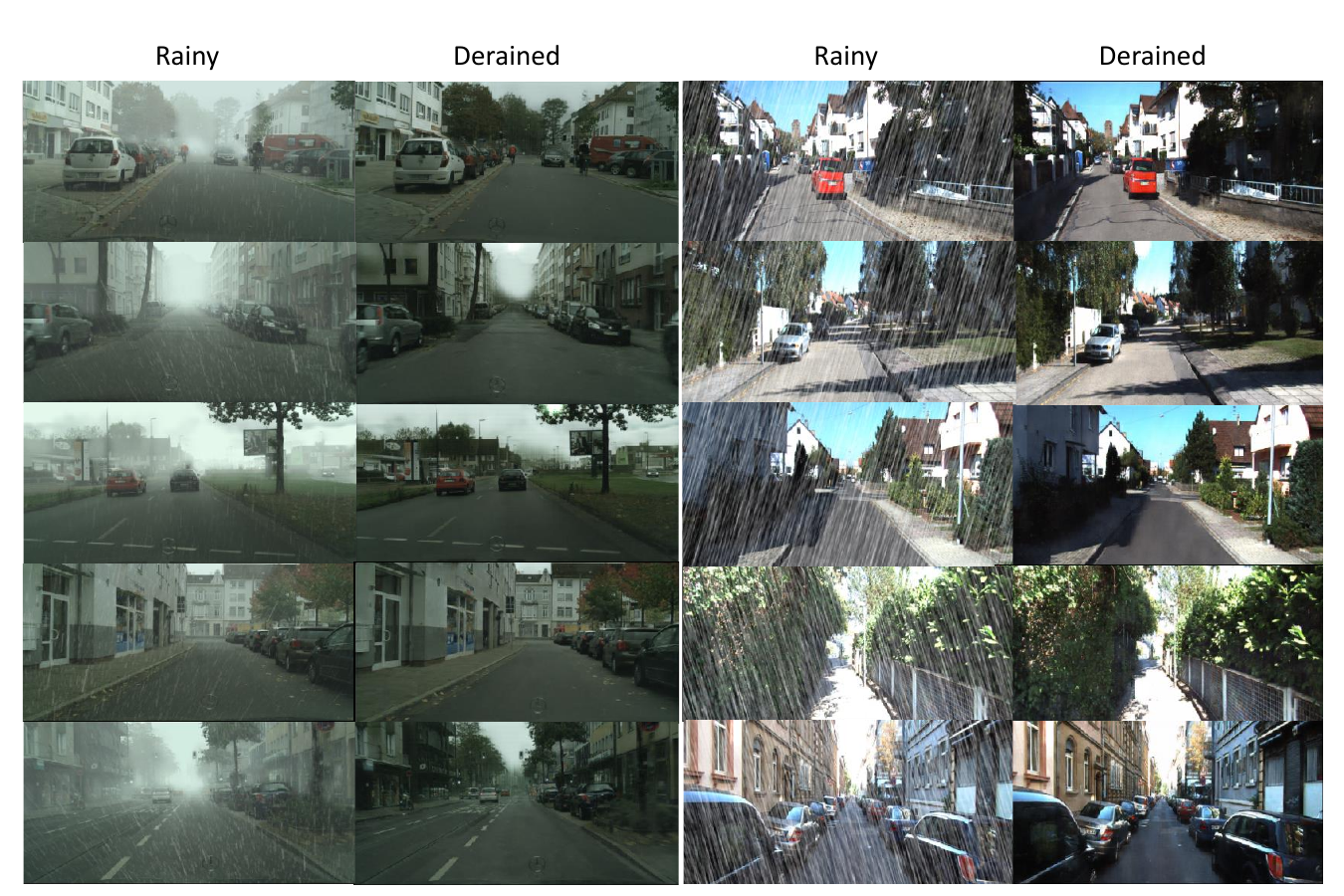}
\caption{Visualization results of RainCityScapes and RainKITTI2012 dataset. The First two columns are exemplar images from the RainCityScapes dataset and corresponding derained outputs; The last two columns are exemplar images from the RainKITTI2012 dataset and corresponding derained outputs.}
\label{VisualResults}
\end{center}
\end{figure}

Table \ref{full_experiment} presents the evaluation metrics for the three datasets. The SSIM demonstrates that our model can restore most of the clear image's information, while the PSNR indicates better overall clarity in the predictions. Figure \ref{VisualResults} shows results of exemplar images from the RainCityScapes and RainKITTI2012 datasets. It is clear that besides rain streaks, the foggy effect has been removed as well.

\begin{table}[ht]
\centering
\caption{We evaluate our model on three outdoor datasets RainCityScapes, RainKITTI2012 and RainKITTI2015, and report the Average (Ave), Maximum (Max), and Minimum (Min) values of PSNR and SSIM.}
\begin{tabular}{c|cccccc}
\hline
\multirow{2}{*}{Datasets} & \multicolumn{3}{c}{PSNR}       & \multicolumn{3}{c}{SSIM}    \\ \cline{2-7} 
                          & Ave      & Max      & Min      & Ave     & Max     & Min     \\ \hline
RainCityScapes            & 28.45243 & 33.58215 & 19.07696 & 0.93726 & 0.97048 & 0.85108 \\
RainKITTI2012             & 25.73460 & 29.70556 & 22.32341 & 0.87256 & 0.92983 & 0.80549 \\
RainKITTI2015             & 26.33563 & 29.74982 & 22.95045 & 0.87402 & 0.91881 & 0.79097 \\ \hline
\end{tabular}

\label{full_experiment}
\end{table}
\subsection{Comparsion with other methods}

We evaluate two additional deraining models, DID-MDN \cite{zhang2018density} and PReNet \cite{ren2019progressive}, on the RainCityscapes testing dataset. For the DID-MDN model, we utilize the pretrained weights provided by the authors on GitHub. Since the DID-MDN model accepts an input size of $512 \times 512$, while RainCityscapes images are sized at $2028 \times 1024$, we resize the RainCityscapes images to $512 \times 512$ for processing, then resize the derained outputs back to the original resolution for evaluation. For PReNet, we leverage all the pretrained models available for Rain100H, Rain100L, and Rain1400 datasets, selecting the best-performing results on the RainCityscapes testing dataset as the final outputs. As seen in the Table \ref{comparsion}, our model can perform better that other methods. Table \ref{performance} presents the inference times of DID-MDN, PReNet, and our method on an NVIDIA RTX A6000 GPU. As shown, our method achieves greater efficiency compared to the other approaches, attributed to its simpler backbone architecture.

\begin{table}[H]
\centering
\caption{Comparsion results on RainCityscapes testing dataset. We report the average, minimum and maximum of PSNR and SSIM mertrics on PReNet, DID-MDN and Our model.}
\begin{tabular}{c|cccccc}
\hline
\multirow{2}{*}{Methods} & \multicolumn{3}{c}{PSNR}       & \multicolumn{3}{c}{SSIM}    \\ \cline{2-7} 
                         & Ave      & Max      & Min      & Ave     & Max     & Min     \\ \hline
DID-MDN                  & 16.82741 & 24.30264 & 11.12157 & 0.77786 & 0.86142 & 0.65167 \\
PRENET                   & 15.75766 & 23.40013 & 10.55631 & 0.80006 & 0.94088 & 0.61976 \\
OURS                     & 28.45243 & 33.58215 & 19.07696 & 0.93726 & 0.97048 & 0.85108 \\ \hline
\end{tabular}

\label{comparsion}
\end{table}

\begin{table}[H]
\centering
\caption{Comparison of inference times on RainCityscapes dataset (image size: 512 x 512).}
\begin{tabular}{c|ccc}
\hline
Method & Inference time \\ \hline
DID-MDN & 0.0322   \\ 
PReNet  & 0.0899 \\ 
Ours & 0.0044 \\ \hline
\end{tabular}
\label{performance}
\end{table}

\subsection{Ablation studies}
All ablation studies are performed on the RainCityscapes, RainKITTI2012, and RainKITTI2015 datasets. To evaluate the effectiveness of our model architecture, we calculate PSNR and SSIM on the respective testing sets. These metrics provide a quantitative assessment of the quality of the generated images, with higher PSNR and SSIM values indicating better image restoration and alignment with ground truth. By comparing different configurations of our model, referred to as Settings A, B, C, D, E, and Full in Table 2, we demonstrate the contributions of each component to the overall performance.
\begin{table}[ht]
\centering
\caption{Ablation settings (A-E). Compared to our full model, we conduct an ablation study by removing component(s) to evaluate their respective contributions.}
\begin{tabular}{c|cccccc}
\hline
Component& A            & B            & C            & D            & E            & Full         \\ \hline
Depth Latent     &              & \checkmark   & \checkmark   & \checkmark   & \checkmark   & \checkmark   \\
Derain Latent    & \checkmark   &              & \checkmark   & \checkmark   & \checkmark   & \checkmark   \\
Ground Truth Depth& \checkmark   & \checkmark   & \checkmark   &              &              & \checkmark   \\
Concatenation of Depth Features& \checkmark   & \checkmark   &              & \checkmark   &              & \checkmark   \\ \hline
\end{tabular}
\label{OverviewAblation}
\end{table}

\textbf{Loss functions} To evaluate the impact of the depth latent and derain latent constraints on our model's performance, we conducted ablation studies on loss components. Table \ref{full_experiment} presents the results of the full model, while Table \ref{wo_depth_latent} and Table \ref{wo_derain_latent} where the depth latent constraint and derain latent constraint are excluded, we observe a noticeable drop in both PSNR and SSIM across all datasets. 

\begin{table}[ht]
\centering
\caption{PSNR and SSIM results of the model trained without depth latent constraint (WO depth latent) on three outdoor datasets: RainCityScapes, RainKITTI2012, and RainKITTI2015.}
\begin{tabular}{c|cccccc}
\hline
Setting A& \multicolumn{3}{c}{PSNR}       & \multicolumn{3}{c}{SSIM}    \\ \hline
Datasets        & Ave      & Max      & Min      & Ave     & Max     & Min     \\ \hline
RainCityScapes  & 25.17939 & 32.74820 & 12.38616 & 0.89550 & 0.95628 & 0.75533 \\
RainKITTI2012   & 25.16171 & 28.75597 & 21.98756 & 0.87104 & 0.92994 & 0.80130 \\
RainKITTI2015   & 25.62023 & 29.53999 & 22.13033 & 0.86560 & 0.91456 & 0.78444 \\ \hline
\end{tabular}
\label{wo_depth_latent}
\end{table}

\begin{table}[ht]
\centering
\caption{PSNR and SSIM results of the model trained without derain latent constraint (WO derain latent) on three outdoor datasets: RainCityScapes, RainKITTI2012, and RainKITTI2015.}
\begin{tabular}{c|cccccc}
\hline
Setting B& \multicolumn{3}{c}{PSNR}       & \multicolumn{3}{c}{SSIM}    \\ \hline
Datasets         & Ave      & Max      & Min      & Ave     & Max     & Min     \\ \hline
RainCityScapes   & 26.49246 & 30.54199 & 21.31583 & 0.92993 & 0.96342 & 0.80553 \\
RainKITTI2012    & 24.80499 & 28.64988 & 21.50745 & 0.86331 & 0.92666 & 0.78119 \\
RainKITTI2015    & 25.56248 & 29.41232 & 22.47581 & 0.87237 & 0.91106 & 0.80080 \\ \hline
\end{tabular}
\label{wo_derain_latent}
\end{table}

\textbf{Ground truth depth} 
Table \ref{wo_gt_depth} shows the performance of the model when trained without using the ground truth depth map (WO GT depth). The results reveal a moderate drop in both PSNR and SSIM across all datasets when the ground truth depth information is removed. 

\begin{table}[H]
\centering
\caption{PSNR and SSIM results of the model trained without the ground truth depth map (WO GT depth) on three outdoor datasets: RainCityScapes, RainKITTI2012, and RainKITTI2015.}
\begin{tabular}{c|cccccc}
\hline
Setting C& \multicolumn{3}{c}{PSNR}       & \multicolumn{3}{c}{SSIM}    \\ \hline
Datasets       & Ave      & Max      & Min      & Ave     & Max     & Min     \\ \hline
RainCityScapes & 27.25449 & 32.11505 & 19.81718 & 0.93005 & 0.96428 & 0.84685 \\
RainKITTI2012  & 24.04377 & 27.61285 & 21.22836 & 0.84602 & 0.91315 & 0.76548 \\
RainKITTI2015  & 25.04490 & 28.21179 & 22.28209 & 0.85778 & 0.90852 & 0.77175 \\ \hline
\end{tabular}
\label{wo_gt_depth}
\end{table}

\textbf{Depth feature concatenation} Table \ref{wo_connection} shows the results of removing depth feature connection between DerainAE encoder and DepthNet encoder, we found that the concatenation of depth features improves the performance.

\begin{table}[H]
\centering
\caption{PSNR and SSIM results of the model trained without depth feature concatenation (WO concatenation) on three outdoor datasets: RainCityScapes, RainKITTI2012, and RainKITTI2015.}
\begin{tabular}{c|cccccc}
\hline
\begin{tabular}[c]{@{}c@{}}Setting D\end{tabular}& \multicolumn{3}{c}{PSNR}       & \multicolumn{3}{c}{SSIM}    \\ \hline
Datasets                                                           & Ave      & Max      & Min      & Ave     & Max     & Min     \\ \hline
RainCityScapes                                                     & 21.09265 & 27.61965 & 15.57796 & 0.84027 & 0.93656 & 0.69801 \\
RainKITTI2012                                                      & 22.04879 & 25.26042 & 19.11878 & 0.79373 &  0.88923 & 0.68814 \\
RainKITTI2015                                                      & 21.74929 & 24.87100 & 19.52019 & 0.81088 & 0.86397 & 0.71841 \\ \hline
\end{tabular}
\label{wo_connection}
\end{table}

\textbf{GT depth and depth feature concatenation}  Table \ref{wo_gt_connection} presents the results when both the ground truth depth map and depth feature concatenation are removed from the model during training. The performance is notably impacted across all datasets, as reflected by the lower PSNR and SSIM values compared to the full model.
\begin{table}[ht]
\centering
\caption{PSNR and SSIM results of the model trained without both ground truth depth map and depth feature concatenation (WO gt depth \& concatenation) on three outdoor datasets: RainCityScapes, RainKITTI2012, and RainKITTI2015.}
\begin{tabular}{c|cccccc}
\hline
\begin{tabular}[c]{@{}c@{}}Setting E\end{tabular}& \multicolumn{3}{c}{PSNR}       & \multicolumn{3}{c}{SSIM}    \\ \hline
Datasets                                                           & Ave      & Max      & Min      & Ave     & Max     & Min     \\ \hline
RainCityScapes                                                     & 24.52006& 32.31414& 14.92033& 0.89442& 0.95443 & 0.74023 \\
RainKITTI2012                                                      & 22.11384& 25.00253& 19.25391& 0.78867& 0.88489 & 0.68008 \\
RainKITTI2015                                                      & 23.69155& 26.68006& 21.29122& 0.82113 & 0.87281 & 0.74432 \\ \hline
\end{tabular}
\label{wo_gt_connection}
\end{table}



\subsection{Vehicle Detection}

\begin{figure}[H]
\begin{center}
\includegraphics[width=1.0\linewidth]{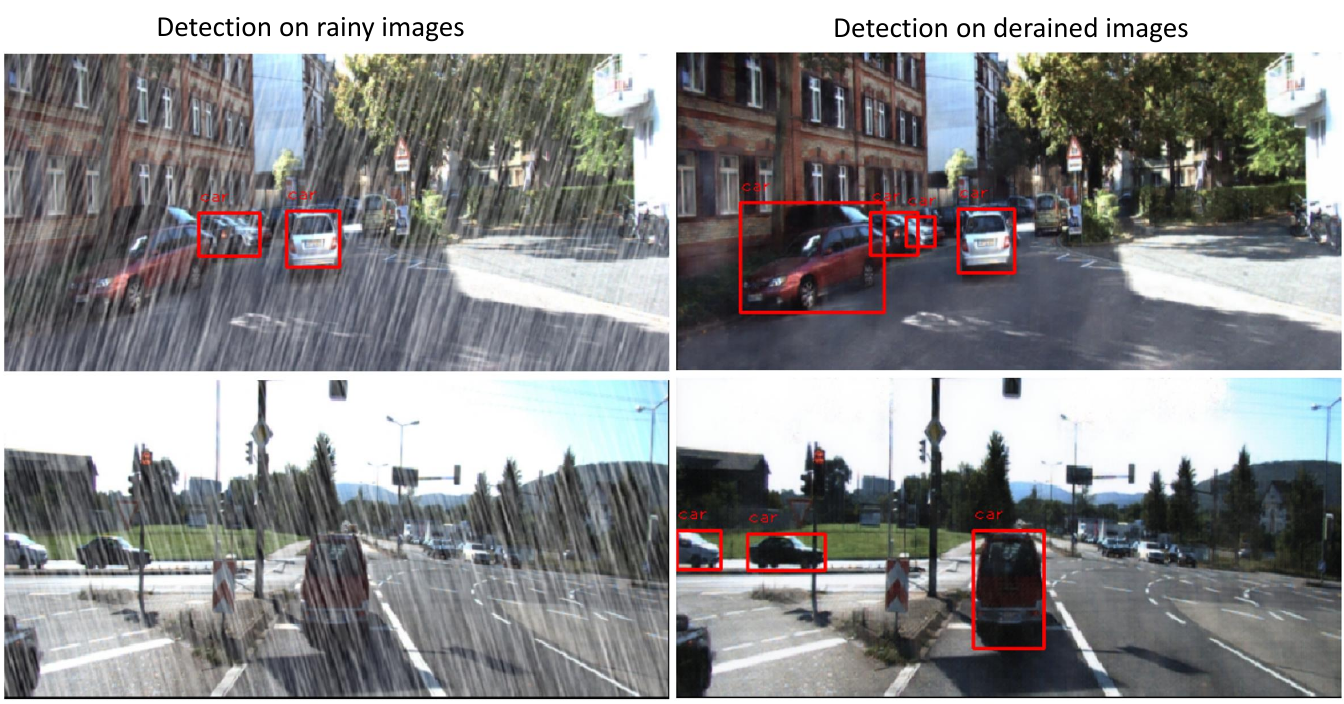}
\caption{Vehicle detection results using YOLOv11 on the RainKITTI2015 dataset.}
\label{Detection}
\end{center}
\end{figure}

\begin{figure}[H]
\begin{center}
\includegraphics[width=1.0\linewidth]{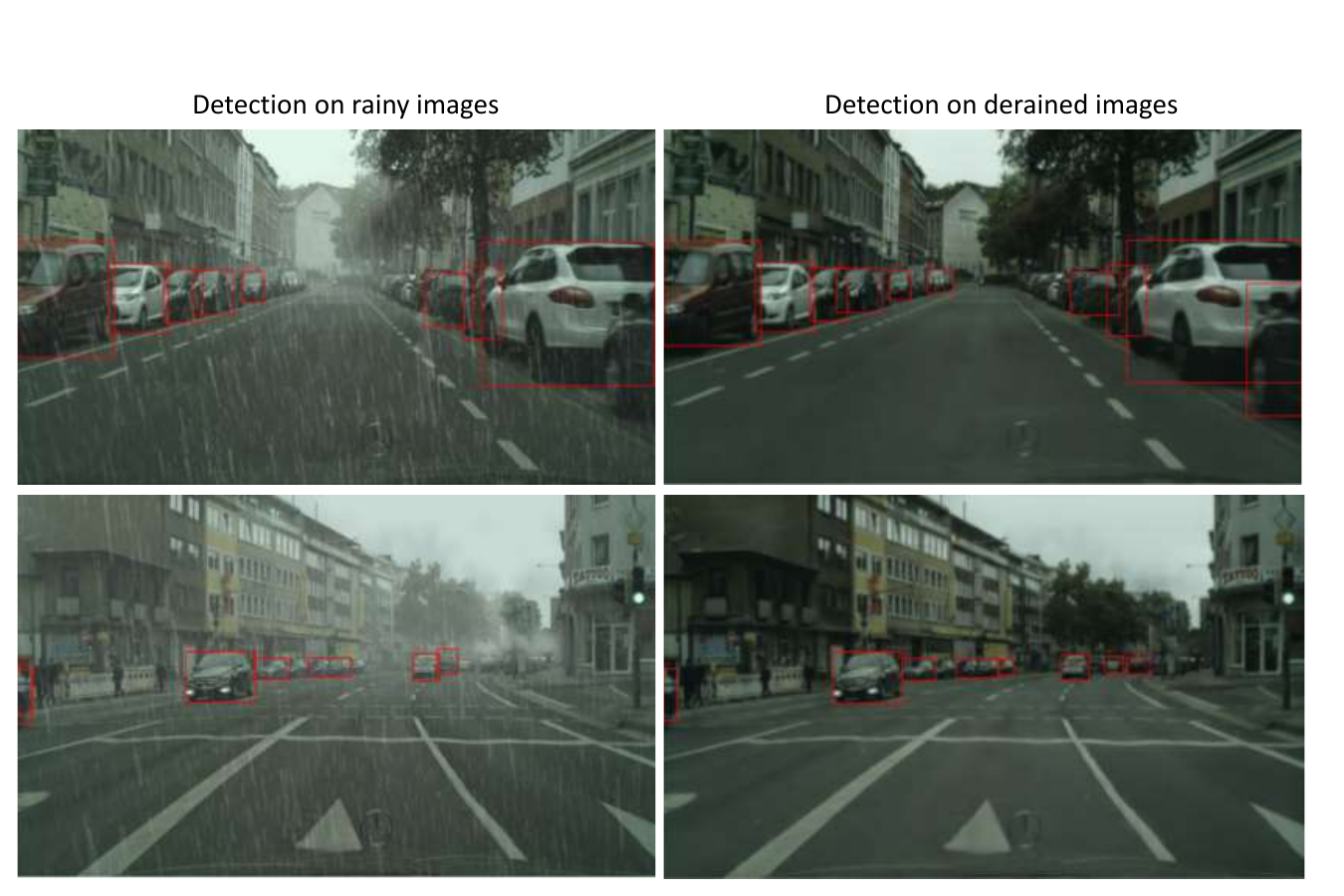}
\caption{Vehicle detection results using YOLOv11 on the RainCityScapes dataset.}
\label{Raincityscapes_detection}
\end{center}
\end{figure}

Image deraining can be integrated into outdoor vision systems to enhance object visibility during complex weather conditions, which is particularly beneficial for autonomous driving. By improving visibility, it can aid in critical tasks like vehicle detection and navigation, making autonomous vehicles safer and more reliable, especially in regions prone to heavy rainfall.  For this evaluation, the focus is on detecting other vehicles in the scene from the ego vehicle perspective. We implemented YOLOv11\cite{khanam2024yolov11} on both rainy and derained images. Figure \ref{Detection} shows that derained images significantly improve vehicle detection accuracy on the RainKITTI2015 dataset. Similarly, Figure \ref{Raincityscapes_detection} demonstrates the ability of our model in enhancing vehicle detection performance under more challenging rainy scenarios in the RainCityscapes dataset, which closely approximate real-world rainy and foggy conditions.

The vehicle detection performance metrics are summarized in Table \ref{car detection}, showing that our deraining model significantly improves vehicle detection recall. This demonstrates enhanced visibility with significantly reduced false negative (missed) detections, which is critical for safe driving of autonomous vehicles, particularly in low-visibility environments.

\begin{table}[H]
\centering
\caption{Vehicle detection results of RainKITTI2015 test dataset. We calculate mean precision and mean recall on 4189 images. The results shows that our deraining model  significantly improves object detection recall.}
\begin{tabular}{c|cc}
\hline
RainKIITI2015 & Mean Precision & Mean Recall \\ \hline
Rainy Image   & 0.9685         & 0.5415      \\
Derain Image  & 0.9533         & 0.9036      \\ \hline
RainCityscapes & Mean precision & Mean recall \\ \hline
Rainy          & 0.823          & 0.628       \\
Derain         & 0.840          & 0.747       \\ \hline
\end{tabular}
\label{car detection}
\end{table}
\section{Conclusions}
In this study, we introduced a novel learning framework that integrates multiple networks, including an AutoEncoder for deraining, an auxiliary network to incorporate depth information, and two supervision networks to enforce feature consistency between rainy and clear scenes. Our approach demonstrates that even with a design based solely on simple convolutional layers, the integration of depth information and feature consistency constraints enables the network to produce high-quality derained images. Our method was evaluated on three public datasets, with results demonstrating its efficacy and robustness under diverse rainy conditions.  Further, applying our model to an object detection task revealed significantly improved recall when using derained images.  It is important to note that the primary focus of this work was not on identifying the optimal model architecture but rather on assessing the impact of different supervisory signals and training strategies.  For future work, we plan to explore more advanced network architectures to further enhance deraining performance, particularly for autonomous driving applications.

\bibliographystyle{unsrt}  
\bibliography{references}  

\end{document}